\documentclass[letterpaper]{article}
\usepackage{aaai}
\usepackage{times}
\usepackage{helvet}
\usepackage{courier}
\usepackage{preamble}
\begin{document}
%
\title{The Weighted Tsetlin Machine:\\Compressed Representations with Weighted Clauses}
\author{Adrian Phoulady$^1$, Ole-Christoffer Granmo$^2$, Saeed R. Gorji$^2$, Hady Ahmady Phoulady$^3$\\
        \small $^1$\href{mailto:adrian.phoulady@gmail.com}{adrian.phoulady@gmail.com}\\
        \small $^2$Centre for Artificial Intelligence Research, University of Agder, Norway\\
        \small $^3$Department of Computer Science, California State University, Sacramento, CA, USA}
\nocopyright
\maketitle
\begin{abstract}
The Tsetlin Machine (TM) is an interpretable mechanism for pattern recognition that constructs conjunctive clauses from data. The clauses capture frequent patterns with high discriminating power, providing increasing expression power with each additional clause. However, the resulting accuracy gain comes at the cost of linear growth in computation time and memory usage. In this paper, we present the Weighted Tsetlin Machine (WTM), which reduces computation time and memory usage by \emph{weighting} the clauses. Real-valued weighting allows one clause to replace multiple, and supports fine-tuning the impact of each clause. Our novel scheme simultaneously learns both the composition of the clauses and their weights. Furthermore, we increase training efficiency by replacing $k$ Bernoulli trials of success probability $p$ with a uniform sample of average size $p k$, the size drawn from a binomial distribution. In our empirical evaluation, the WTM achieved the same accuracy as the TM on MNIST, IMDb, and Connect-4, requiring only $1/4$, $1/3$, and $1/50$ of the clauses, respectively. With the same number of clauses, the WTM outperformed the TM, obtaining peak test accuracies of respectively $98.63\%$, $90.37\%$, and $87.91\%$. Finally, our novel sampling scheme reduced sample generation time by a factor of $7$.
\end{abstract}

\maketitle
%
%
\keywords{Tsetlin Machines, Pattern Recognition, Propositional Logic, Learning Automata, Frequent Pattern Mining.}
%
\section{Introduction}
The Tsetlin Machine (TM) is a novel learning mechanism that was introduced in 2018 \cite{granmo2018tsetlin}. Like artificial neural networks, the TM can map an exponential number of feature value combinations to an appropriate response. However, neural networks are based on multiplication, accumulation, and non-linear activation across multiple layers of neurons, whereas the TM is based on propositional logic formulas, formed by learning automata. 

For $n$ binary features, there can be $2^{2^n}$ propositional formulas. Equipped with an adequate number of conjunctive clauses, a TM can learn any of them in a relatively simple form. In addition to being computationally simple, the models built by the TM have the advantage of being interpretable by human experts, which may be of vital importance in sensitive applications such as medical decision making~\cite{berge2019,Rudin2019}.

\paragraph{Learning Automata.}
Learning automata have attracted considerable interest because they can learn the optimal action when operating in unknown stochastic environments \cite{thathachar1987learning}. As such, they have been used for pattern classification over more than four decades. Early work includes stochastic learning automata based classifiers and games of learning automata \cite{barto1985pattern,thathachar1987learning}. These approaches can learn the optimal classifier for specific forms of discriminant functions (e.g., linear classifiers) also when feedback is noisy. Along the same lines, Sastry and Thathachar have provided several algorithms based on cooperating systems of learning automata \cite{sastry1999learning}. 

More recently, Zahiri proposed hyperplane- and rule-based classifiers, which performed comparably to other well-known methods in the literature \cite{zahiri2008learning,zahiri2012classification}. Recent research also includes a noise-tolerant learning algorithm built upon a team of continuous-action learning automata \cite{sastry2009team}. Further, Goodwin, Yazidi, and Jonassen proposed a learning automata-guided random walk on a grid to construct robust discriminant functions \cite{goodwin2016distributed}. Finally, Motieghader et al. introduced a hybrid scheme that combines a genetic algorithm with learning automata to address the gene selection problem in cancer classification \cite{motieghader2017hybrid}. 

In general, previous learning automata schemes have mainly addressed relatively small-scale pattern recognition tasks. Lately, however, several TM-based approaches have demonstrated promising scalability properties. This includes the natural language understanding approach by Berge et al., which uses the conjunctive clauses of the TM to capture textual patterns \cite{berge2019}. Also, TM-based convolution for image analysis was introduced by Granmo et al., employing the clauses as interpretable filters \cite{granmo2019convtsetlin}. Further, Abeyrathna et al. designed a novel TM for regression \cite{abeyrathna2019regressiontsetlin} that supports continuous input \cite{abeyrathna2019scheme}. All of these schemes compare favourably with state-of-the-art pattern classification and regression techniques.


\paragraph{Paper Contributions.} In this paper, we introduce the Weighted Tsetlin Machine (WTM), which employs \emph{weighted conjunctive clauses}. Our intent is to significantly reduce memory usage and computation time by using the clause weights to compress the pattern representation of the TM. We start with giving a brief overview of the TM in Section~\ref{sec:tsetlin_machine}. Then, in Section~\ref{sec:weighted_tsetlin_machine}, we introduce the WTM and the concept of weighted clauses, that is, clauses that output continuous rather than binary values. Real-valued weighting allows one clause to replace multiple, and supports fine-tuning the impact of each clause. Our novel scheme simultaneously learns both the composition of each clause as well as their weights. Further, we simplify sample generation during learning by replacing Bernoulli process sampling with sampling from a binomial distribution followed by uniform sampling. We evaluate our approach empirically on MNIST, IMDb, and Connect-4 in Section \ref{sec:results}, demonstrating that the WTM outperforms the TM in all the experiments. We conclude in Section~\ref{sec:conclusion}, and present paths ahead for further research.

\section{The Tsetlin Machine}
\label{sec:tsetlin_machine}
TM learning is based on forming formulas in propositional logic to capture frequent patterns of high discriminating power. We here describe how the TM operates multiple teams of so-called Tsetlin automata \cite{tsetlin1961behaviour} to build discriminative conjunctive clauses, followed by majority voting to decide the final classification from the clause outputs.
\subsection{The Tsetlin Automaton}
\begin{figure*}[ht]
\centering
\colorlet{rcolor}{darkgreen}
\colorlet{pcolor}{darkerred}
\colorlet{scolor}{darkergray}
\begin{tikzpicture}[>=stealth, shorten >=0pt,node distance=1.5cm]
  \node [state, scolor] (S1)                {$1$};
  \node [state, scolor] (S2)  [right of=S1] {$2$};
  \node [scolor]        (D1)  [right of=S2] {$\cdot\cdot\cdot$};
  \node [state, scolor] (SN)  [right of=D1] {$N$};
  \node [state, scolor] (SN1) [right of=SN,inner sep=0] {$N{+}1$};
  \node [state, scolor] (SN2) [right of=SN1,inner sep=0] {$N{+}2$};
  \node [scolor]        (D2)  [right of=SN2] {$\cdot\cdot\cdot$};
  \node [state, scolor] (S2N) [right of=D2] {$2N$};
  
	\path[->] (S1) edge [loop left, looseness=8, rcolor]    node               {$\beta_2$} (S1);
	\path[->] (S2) edge [bend right, rcolor] node [above]  {$\beta_2$} (S1);
	\path[->] (D1) edge [tail=.6,bend right, rcolor] (S2);
	\path[-] (SN) edge [head=.6,bend right, rcolor] (D1);
	\path[->] (S1) edge [densely dashed, pcolor, bend right] node [below]  {$\beta_1$} (S2);
	\path[-] (S2) edge [head=.6, densely dashed, pcolor, bend right] (D1);
	\path[->] (D1) edge [tail=.6, densely dashed, pcolor, bend right] (SN);

	\path[->] (S2N) edge [loop right, looseness=8, rcolor]    node               {$\beta_2$} (S2N);
	\path[<-] (SN2) edge [bend right, rcolor] node [above]  {$\beta_2$} (SN1);
	\path[-] (SN2) edge [head=.6,bend left, rcolor] (D2);
	\path[->] (D2) edge [tail=.6,bend left, rcolor] (S2N);
	\path[<-] (SN1) edge [densely dashed, pcolor, bend right] node [below]  {$\beta_1$} (SN2);
	\path[->] (D2) edge [tail=.6, densely dashed, pcolor, bend left] (SN2);
	\path[-] (D2) edge [tail=.6, densely dashed, pcolor, bend right] (S2N);
	\path[->] ([yshift=+1mm] SN.east) edge [densely dashed, pcolor] node [above]  {$\beta_1$} ([yshift=+1mm] SN1.west);
	\path[<-] ([yshift=-1mm] SN.east) edge [densely dashed, pcolor] node [below]  {$\beta_1$} ([yshift=-1mm] SN1.west);

	\path[-, thick, gray] ([yshift=-11mm] S1.west) edge node [fill=white] {$\alpha_1$} ([yshift=-11mm] SN.east);
	\path[-, thick, darkergray] ([yshift=-11mm] SN1.west) edge node [fill=white] {$\alpha_2$} ([yshift=-11mm] S2N.east);
\end{tikzpicture}
\caption{A two-action Tsetlin automaton with $2N$ internal states}\label{fig:tsetlin_automaton}
\end{figure*}
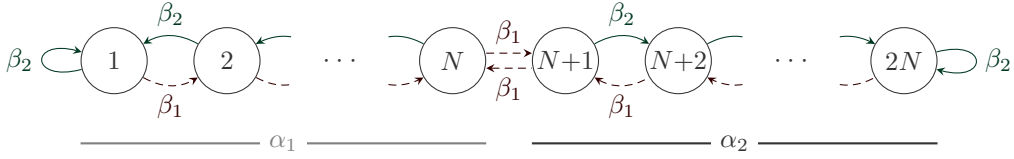

A two-action {\em Tsetlin automaton} \cite{tsetlin1961behaviour,narendra2012learning} is a learning automaton with $2N$ states $\phi_1$, $\phi_2$, \ldots, $\phi_{2N}$ and two actions $\alpha_1$, $\alpha_2$ (Figure~\ref{fig:tsetlin_automaton}). It performs its actions sequentially in an environment. That is, in the states $\phi_1$, \ldots, $\phi_N$, the Tsetlin automaton performs action $\alpha_1$, and in the states $\phi_{N+1}$, \ldots, $\phi_{2N}$, it performs action $\alpha_2$. The environment responds with two types of input to the automaton: penalties $\beta_1$ and rewards $\beta_2$. A state transition function governs learning. With the penalty input $\beta_1$, the transition function changes the state $\phi_i$ to $\phi_{i+1}$ for $1\le i\le N$ and to $\phi_{i-1}$ for $N+1\le i\le2N$. Conversely, the reward input $\beta_2$ makes the state $\phi_i$ change to $\phi_{i-1}$ for $1<i\le N$ and to $\phi_{i+1}$ for $N+1\le i<2N$, while leaving it unchanged for $i=1,2N$. In effect, $\beta_1$ makes the Tsetlin automaton change state toward the centre, whereas $\beta_2$ moves the state away from the centre, towards the extreme ends $1$ or $2N$.

\subsection{Tsetlin Machine Structure and Inference}
A binary classifier over $o$ binary features can be regarded as a Boolean function --- each of the $2^o$ possible combinations of the $o$ binary inputs belongs to one of the classes $0$ or $1$.

A Boolean function can always be represented as a disjunction of conjunctive clauses, referred to as disjunctive normal form. For binary input features $\mathbf x=(x_1,\ldots,x_o)$, each conjunctive clause $C_j$ is represented by a subset $L_j$ of the literal set $L=\{x_1, \ldots, x_o,\lnot x_1, \ldots, \lnot x_o\}$. Knowing $L_j$, we have
\begin{equation}
C_j(\mathbf x)=\bigwedge_{l\in L_j}l.
\end{equation}
The Boolean function $f$ in disjunctive normal form with clauses $C_1$, \ldots, $C_m$ then becomes
\begin{align}
f(\mathbf x)=\bigvee_{\vphantom{l}j=1}^m C_j(\mathbf x)=\bigvee_{\vphantom{l}j=1}^m\bigwedge_{l\in L_j}l.
\end{align}

Similar to Karnaugh maps \cite{karnaugh1953map}, conjunctive clauses form the core of function representation in the TM. In brief, a TM learns the composition of each conjunctive clause by specifying its literals using a team of $2o$ Tsetlin automata, one Tsetlin automaton per literal. Each Tsetlin automaton decides between two actions. Action $\alpha_1$ means excluding the associated literal, whereas action $\alpha_2$ means including it in the conjunctive clause. To exemplify, Figure~\ref{fig:tsetlin_automata_team} shows the configuration of a team of eight Tsetlin automata that form a conjunctive clause from the three binary features $x_1$, $x_2$, and $x_3$.

\begin{figure}[ht]
\centering
\definecolor{acolor}{rgb}{0.0, 0.1, 0.2}
\colorlet{icolor}{darkgreen}
\colorlet{ecolor}{darkerred}
\colorlet{acolor}{darkergray}
\tikzset{automaton/.style = {rectangle, draw, acolor, minimum width=.9cm, minimum height=.7cm},
exclude/.style = {densely dashed, ecolor, opacity=0.8},
include/.style = {icolor}}
\begin{tikzpicture}[>=stealth, shorten >=0pt,node distance=1.2cm]
  \node [automaton]                      (TA1) {$\text{TA}_1$};
  \node [automaton, right of=TA1] (TA2) {$\text{TA}_2$};
  \node [automaton, right of=TA2] (TA3) {$\text{TA}_3$};
  \node [automaton, right of=TA3] (TA4) {$\text{TA}_4$};
  \node [automaton, right of=TA4] (TA5) {$\text{TA}_5$};
  \node [automaton, right of=TA5] (TA6) {$\text{TA}_6$};
  \path[->] (TA1) edge [include] node [right]  {$\alpha_2$}
        node [above=.3cm, black] {$x_1$}
        +(0, .8) node [below=.5cm] {$x_1$};
  \path[->] (TA2) edge [exclude] node [right]  {$\alpha_1$}
        +(0, .8) node [below=.5cm] {$x_2$};
  \path[->] (TA3) edge [include] node [right]  {$\alpha_2$}
        node [above=.3cm, black] {$x_3$}
        +(0, .8) node [below=.5cm] {$x_3$};
  \path[->] (TA4) edge [exclude] node [right]  {$\alpha_1$}
        +(0, .8) node [below=.5cm] {$\lnot x_1$};
  \path[->] (TA5) edge [include] node [right]  {$\alpha_2$}
        node [above=.3cm, black] {$\lnot x_2$}
        +(0, .8) node [below=.5cm] {$\lnot x_2$};
  \path[->] (TA6) edge [exclude] node [right]  {$\alpha_1$}
        +(0, .8) node [below=.5cm] {$\lnot x_3$};
\end{tikzpicture}
\caption{A team of six two-action Tsetlin automata forming the clause $x_1\land x_3\land\lnot x_2$}
\label{fig:tsetlin_automata_team}
\end{figure}

Instead of forming a disjunction of the conjunctive clauses, a TM sums up the clause outputs to produce an ensemble effect. While any propositional formula can be represented in disjunctive normal form, it turns out that the introduced ensemble effect helps dealing with noisy data \cite{granmo2018tsetlin}. Further, the TM groups the clauses into positive ones $C^+_1$, $C^+_2$, \ldots, $C^+_{c_P}$ and negative ones $C^-_1$, $C^-_2$, \ldots, $C^-_{c_N}$. In effect, for the input $\mathbf x=(x_1,\ldots,x_o)$, the TM computes the signed sum
\begin{align}
s(\mathbf x)=\sum_{j=1}^{c_P}C^+_j(\mathbf x)-\sum_{j=1}^{c_N}C^-_j(\mathbf x)
\label{eqn:s(x)}
\end{align}
to perform classification. If the signed sum is negative, the TM classifies the input to class $\hat y=0$, and otherwise, it classifies it to class $\hat y=1$. In other words, the classification is performed by applying the unit step function on $s(\mathbf x)$:
\begin{align}
\hat y=u\bigl(s(\mathbf x)\bigr)=u\biggl(\sum_{j=1}^{c_P}C^+_j(\mathbf x)-\sum_{j=1}^{c_N}C^-_j(\mathbf x)\biggr).
\label{eqn:y_hat}
\end{align}

\subsection{Tsetlin Machine Learning}
The TM builds upon reinforcement learning, with feedback given directly to the conjunctive clauses. Each clause, in turn, passes the feedback onward to its individual Tsetlin automata. There are two types of feedback: Feedback Type~I and Feedback Type~II. When a clause receives Type~I feedback, its automata are modified so that the clause eventually evaluates to $1$ for the current input. If it already evaluates to $1$, Type~I feedback will instead refine the clause by inserting more literals. Type~II feedback, on the other hand, makes the affected clauses eventually evaluate to $0$. In combination, these two types of feedback gradually modify which literals are present in each clause in a way that over time improves the classification accuracy.

\subsubsection{Overall Feedback Loop}
A TM is initialized by randomly setting the states of the individual Tsetlin automata, thus producing clauses with randomly selected literals. From there, training data is fed to the TM, one example $(\mathbf x, y)$ at a time. In other words, the learning can be performed on-line. For an input $\mathbf x$ of class $y=0$, the goal is to make the signed sum $s(\mathbf x)$ of the clauses become negative (see Equation~\ref{eqn:y_hat}). If this is not the case, we mitigate by giving some of the negative clauses Type~I feedback, and some of the positive clauses Type~II feedback. For input $\mathbf x$ of class $y=1$, on the other hand, the learning goal is to make the signed sum $s(\mathbf x)$ become non-negative (again see Equation~\ref{eqn:y_hat}). Accordingly, if the signed sum of the clauses is negative, we increase it by giving some of the positive clauses Type~I feedback and some of the negative clauses Type~II feedback.

To introduce an ensemble effect, feedback is given to a random selection of the clauses based on a hyperparameter $T$ -- the target of summation. In brief, the TM strives to make the signed sum $s(\mathbf x)$ reach $-T$ for an input of class $y=0$. Conversely, it seeks to make $s(\mathbf x)$ reach $T$ for an input of class $y=1$. To achieve this, we first clamp $s(\mathbf x)$ between $-T$ and $T$: $c(\mathbf x)=\text{clamp}\bigl(s(\mathbf x), -T, T\bigr)$. Second, each clause receives feedback with probability $p_c(\mathbf x)$ proportional to the difference between the clamped sum and the summation target:
\begin{align}
p_c(\mathbf x)=
	\begin{cases}
		\displaystyle\frac{T+c(\mathbf x)}{2T},&\quad\text{if } y=0,\\[1.5ex]
		\displaystyle\frac{T-c(\mathbf x)}{2T},&\quad\text{if } y=1.
	\end{cases}
\end{align}
The above random clause selection has an additional crucial effect: it guides the clauses to distribute themselves across the significant frequent sub-patterns, as opposed to clustering on a few ones. This is achieved by making the expected frequency of feedback to each clause proportional to the difference between the clamped sum $c(\mathbf x)$ and the summation target $\pm T$. That is, feedback activity calms down as $s(\mathbf x)$ approaches its target $\pm T$, and comes to a complete standstill when the target is met or surpassed. In this manner, only a fraction of the available clauses are stimulated to recognize each frequent sub-pattern.

\subsubsection{Type~I Feedback}
We first cover the details of Type~I feedback, which plays two roles. Firstly, Type~I feedback attempts to modify clauses that evaluate to $0$ for the current input $\mathbf{x}$, so that they eventually start outputting $1$ instead. This is achieved by making the clauses \emph{sparser}, removing literals one by one. If a clause already outputs $1$, however, Type~I feedback refines the clause by including more literals, making it \emph{denser}.

\begin{table*}[t]
\centering
\begin{tabular}{l*{4}{p{5.2em}}}
\toprule
\bfseries Clause value&\multicolumn{2}{c}{\bfseries0}&\multicolumn{2}{c}{\bfseries1}\\[1ex]
\bfseries Literal value&\hfil\bfseries0&\hfil\bfseries1&\hfil\bfseries0&\hfil\bfseries1\\
\midrule[\heavyrulewidth]
\bfseries Inclusion response&\hfil$p_s$-penalty&\hfil$p_s$-penalty&\hfil NA&\hfil reward\\[1ex]
\bfseries Exclusion response&\hfil$p_s$-reward&\hfil$p_s$-reward&\hfil$p_s$-reward&\hfil penalty\\
\bottomrule
\end{tabular}
\caption{Type~I Feedback}
\label{tab:type_i}
\end{table*}

Table~\ref{tab:type_i} contains the rules by which Type~I feedback operates. Recall that each Tsetlin automaton is associated with a specific literal, within a specific clause. As seen in the table, the value of the clause in combination with the value of the literal governs feedback to the associated Tsetlin automaton.
\begin{itemize}
\item[(1)] If both the clause and the literal are of value $1$, the Tsetlin automaton receives feedback that stimulates including the literal. The purpose is to refine the encompassing clause, so that it provides a more specific representation of the input $\bf x$.
\item[(2)] All the other Tsetlin automata (those associated with clauses or literals of value $0$), on the other hand, receive feedback that stimulates excluding their literals. This occurs randomly with probability $p_s$. The goal here is to combat overfitting by making the clause sparser, and in the extreme, erase the pattern completely to make room for a new.
\end{itemize}
Note that the probability $p_s$ is a hyperparameter that controls the sparsity of the clauses produced. In brief, the sparsity of clauses increases with $p_s$, simply because a higher $p_s$ produces more exclude actions relative to include actions.

Also note that as learning proceeds, Type I feedback makes the states of the Tsetlin automata gradually move away from the centre states. The automata thus become increasingly confident in their decisions. This stabilizes the composition of the clauses, calming down exploration.

\subsubsection{Type~II Feedback}
The purpose of giving Type~II feedback to a clause is to ensure that it eventually outputs $0$ for the current input. Table~\ref{tab:type_ii} contains the rules for Type~II feedback. As seen, it is only when a clause outputs $1$ that its Tsetlin automata are given Type~II feedback, and then only those automata whose literals are of value $0$. These Tsetlin automata are quite simply penalized for their exclude actions. Accordingly, sooner or later, one of the excluded literals are included instead, turning the clause output to $0$.

Contrary to Type~I feedback, Type~II feedback is not intended to make the Tsetlin automata more confident, or the clauses denser/sparser. Rather, Type~II feedback disrupts the targeted clause by introducing new literals that potentially have high discrimination power. These literals are only picked up by Type~I feedback if they take part in frequent sub-patterns. Otherwise, Type~I feedback will throw them back into exclusion. In this manner, overfitting is combated, while discrimination power is increased.

\begin{table*}[t]
\centering
\begin{tabular}{l*{4}{p{5.2em}}}
\toprule
\bfseries Clause value&\multicolumn{2}{c}{\bfseries0}&\multicolumn{2}{c}{\bfseries1}\\[1ex]
\bfseries Literal value&\hfil\bfseries0&\hfil\bfseries1&\hfil\bfseries0&\hfil1\\
\midrule[\heavyrulewidth]
\bfseries Inclusion response&\hfil--&\hfil--&\hfil NA&\hfil--\\[1ex]
\bfseries Exclusion response&\hfil--&\hfil--&\hfil penalty&\hfil--\\
\bottomrule
\end{tabular}
\caption{Type~II Feedback}
\label{tab:type_ii}
\end{table*}

\section{The Weighted Tsetlin Machine}
\label{sec:weighted_tsetlin_machine}
We now introduce the Weighted Tsetlin Machine (WTM), which enhances the classic TM with weighted clauses as well as improving sampling efficiency.\footnote{The code for the Weighted Tsetlin Machine can be found at \url{https://github.com/adrianphoulady/weighted-tsetlin-machine-cpp} (C++), and \url{https://github.com/cair/pyTsetlinMachine} (Python/C).}

\subsection{Binomial Distribution Based Type~I Feedback}
Random number generation is relatively time consuming, yet an essential part of giving Type~I feedback. That is, when randomizing feedback to the $2o$ Tsetlin automata of a clause, one samples from a Bernoulli process of length $u=2o$ with success probability $p_s$.

We here propose to increase training efficiency by leveraging that $p_s$ typically is small in TM learning, and that the number of successes $S_u$ of the Bernoulli process has a Binomial distribution:
\begin{align}
P(S_u=k)=\binom u k p_s^k(1-p_s)^{u-k}.
\end{align}
To single out the Tsetlin automata to receive feedback more efficiently, we first assume that there are no successes: $B[1\twodots u] \gets 0$. We then  obtain the number of successes $q$ by sampling once from the corresponding binomial distribution. Finally, we update $B[1\twodots u]$ by randomly selecting $q$ Tsetlin automata.
\begin{algorithm}[ht]
\begin{algorithmbox}
\Procname{$\proc{Binomial-Uniform-Sampling}(u,p)$}
\li $B[1\twodots u] \gets 0$
\li $\id{q} \gets \proc{Binomial}(u,p)$
\li $\id{k} \gets 1$
\li \While $k \le q$
\li \Do
	$\id{v} \gets \proc{Uniform}(1,u)$
\li	\If $B[v] \neq 1$
\li	\Do
		$B[v] \gets 1$
\li		$k \gets k + 1$
	\End
  \End
\li \Return $B$
\end{algorithmbox}
\caption{Generating $u$ Bernoulli process samples with success probability $p$ from a binomial distribution}
\label{alg:bernoullib}
\end{algorithm}
As summarized in Algorithm~\ref{alg:bernoullib}, we thus produce the complete sample by generating about $2p_so+1$ random values on average, rather than $2o$.

\subsection{Weighted Clauses}
Due to the ensemble effect of TM learning, several versions of similar clauses may occur multiple times in the final classifier. The question we investigate here is whether the collection of clauses as a whole can be represented more compactly by \emph{weighting} the clauses. Then, instead of repeating a clause $w$ times, we can keep a single instance, assigning it the weight $w$. Further, by allowing $w$ to be real-valued, we can introduce \emph{fractional} clauses. Thus, if we want to halve the effect of a clause, we simply give it the weight $w=0.5$. In this manner, a more compact representation can be obtained by using fewer clauses, reducing the number of model parameters, to facilitate both faster learning and faster classification. 

In all brevity, we associate a positive real-valued weight $w_j$ with each clause $j$, multiplying the weights with the respective clause outputs: \begin{align}
s'(\mathbf x)=\sum_{j=1}^{c_P}w^+_j C^+_j(\mathbf x)-\sum_{j=1}^{c_N}w^-_jC^-_j(\mathbf x).
\end{align}
In this structure, to increase or reduce the impact of a clause, one merely needs to adjust its weight. Furthermore, the overall sum $s'(\mathbf x)$ is now real-valued, generalizing the TM.

\subsection{WTM Recognition}
Recognition in the WTM is conducted in the same way as for the TM. That is, we simply apply the step function to the weighted sum:
\begin{align}
\hat y=u\bigl(s'(\mathbf x)\bigr)=u\biggl(
\sum_{j=1}^{c_P}w^+_j C^+_j(\mathbf x)-\sum_{j=1}^{c_N}w^-_jC^-_j(\mathbf x)\biggr).
\end{align}
The multiclass WTM operates similarly, too. In brief, with $n$ classes, we introduce one WTM per class $i=1,\ldots,n$. The resulting weighted sums $s'_i(\mathbf x)$ are then used for  classification. That is, the largest weighted sum determines the predicted class:
\begin{align}
\hat y=\mathop{\mathrm{argmax}}_i\{s'_i(\mathbf x)\}.
\end{align}

\subsection{WTM Learning}
The crucial question that remains is how the weights can be learned to provide a compact, yet accurate pattern representation. We here propose a simple scheme for learning the weights.

\paragraph{Initialization.} First, we initialize all the weights to $1.0$:
\begin{align}
w^+_j&\gets 1.0,\\
w^-_j&\gets 1.0.
\end{align}
Thus, at the start of learning, the behaviour of the clauses are identical to those of the TM.

\paragraph{Weight Updating.} Adjusting the Tsetlin automata's states in the WTM is performed in the same way as for the TM. However, we introduce a novel approach for updating the weights. This updating is governed by Type~I and Type~II feedback. Note that, below, we only update the weights of those clauses that evaluate to~$1$ for the current input $\mathbf x$.

\paragraph{Learning Rate.} Weight learning is controlled by a \emph{learning rate} $\gamma\in[0,\infty)$. For Type~I feedback, each clause $C^+_j$/$C^-_j$ outputting $1$ has its weight $w^+_j$/$w^-_j$ updated by multiplying it by $1+\gamma$:
\begin{align}
w^+_j&\gets w^+_j\cdot(1+\gamma),\quad\text{if }C^+_j(\mathbf x)=1,\\
w^-_j&\gets w^-_j\cdot(1+\gamma),\quad\text{if }C^-_j(\mathbf x)=1.
\end{align}
For Type~II feedback, we instead update the weights by dividing by $1+\gamma$:
\begin{align}
w^+_j&\gets w^+_j\,/\,(1+\gamma),\quad\text{if }C^+_j(\mathbf x)=1,\\
w^-_j&\gets w^-_j\,/\,(1+\gamma),\quad\text{if }C^-_j(\mathbf x)=1.
\end{align}
When a clause evaluates to $0$, its weight remains unchanged:
\begin{align}
w^+_j&\gets w^+_j,\quad\text{if }C^+_j(\mathbf x)=0,\\
w^-_j&\gets w^-_j,\quad\text{if }C^-_j(\mathbf x)=0.
\end{align}
As seen, for Type~I feedback, the weights are increased to strengthen the impact of the associated clauses, thus reinforcing true positive frequent patterns. For Type~II feedback, on the other hand, the weights are decreased instead. This is to diminish the impact of the associated clauses, thus combating false positives. Observe that for $\gamma=0$, the weights of the clauses are fixed at the initial value $1.0$, making the resulting WTM equivalent to a TM.


\def\nothing{
\subsection{Coefficient Normalization}
In TM, there is a threshold $T$, and training tries to push the signed sum out of the interval $(-T,T)$. If the coefficients of negative and positive clauses are doubled, the signed sum will be doubled. This helps the signed sum to get larger in the magnitude. Thus, the threshold encourages the coefficient to get larger, not on the necessity of better classification but for filling the gap between the signed sum and the boundaries. This phenomenon can be avoided by coefficient normalization.

The initial sum of the coefficients is equal to $c_P+c_N$, the total number of clauses. After training for each sample, the coefficients may change. We multiply every coefficient $f_i$ by the normalization factor
\begin{align}
\nu=\frac{c_P+c_N}{\sum_{i=1}^{c_P}f^+_i+\sum_{j=1}^{c_N}f^-_j}
\end{align}
so that the sum of the coefficients remains the same:
\begin{align}
f_i^{\mathit{new}}=\nu f_i.
\end{align}
}

\section{Empirical Evaluation} \label{sec:results}
We here present empirical results on the performance of the WTM in comparison with the TM. To this end, we use datasets from three different areas of pattern recognition: MNIST handwritten digit recognition \cite{lecun1998gradient}, IMDb sentiment analysis \cite{maas-EtAl:2011:ACL-HLT2011}, and Connect-4 game winner prediction \cite{Dua:2019}. We further investigate the behaviour of the novel WTM mechanisms in more detail, including learning speed and weight distribution.

\subsection{MNIST Handwritten Digit Recognition}
The MNIST dataset contains images of handwritten digits, $60{,}000$ training and $10{,}000$ test examples. Each example is a labelled $28\times28$-pixel grayscale image 
The pixels are represented by integers between $0$ (black) and $255$ (white), and each label is an integer from $0$ to $9$. We binarize the input pixels to get $768$ bits of data. A value less than $77=\lceil.3\cdot255\rceil$ is converted to~$0$, and a value greater than or equal to $77$ is converted to~$1$. Note that other binarization thresholds provided similar performance because the image pixels are mainly black and white.

\subsubsection{Feedback Generation Speedup}
Employing $2{,}000$ class-clauses on MNIST produces $12.6$ millions Type~I feedback calls in the first epoch
.
In turn, each call samples $n=2\cdot28\cdot28=1568$ values from a Bernoulli process with success probability $p_s=.1$, to assign feedback to the individual Tsetlin automata. This is the most time consuming part of TM learning. Whereas recognizing the $60{,}000$ images takes $23.5$ seconds, feedback generation using standard Bernoulli process sampling takes $61.2$ seconds. This is about $2/3$ of the training time in the first epoch. However, replacing the Bernoulli process sampling with our binomial distribution based sampling approach (Algorithm~\ref{alg:bernoullib}), we are able to speed up the random feedback generation by a factor of $7$, significantly improving the overall learning speed of the TM.
 

\subsubsection{Weighted Clauses with WTM}
We now turn to evaluating the effect of weighted clauses. In our first experiment we used a $10$-class TM with $2{,}000$ clauses per class as baseline. The accuracy obtained by this configuration was parred by a WTM with merely $500$ clauses. This reduced  memory usage $4$ times, while increasing execution time by a factor of $4$. Together with the fast feedback generation procedure, the full WTM ran
approximately $10$ times faster than the TM, measured over $300$ epochs of training.

\begin{table}[!ht]
\centering
\tableit
{98.24}{97.97}{98.11}{0.07}{98.02}{98.23}
{98.63}{98.39}{98.50}{0.05}{98.43}{98.60}
\caption{MNIST test accuracy statistics of the TM and the WTM with $4{,}000$ clauses per class}
\label{tab:mnist-acc}
\end{table}

Furthermore, by running the TM and WTM with the same number of clauses ($4{,}000$ per class) we observed that the WTM was significantly more accurate than the TM, with slightly faster execution time (1.7 minutes versus 2 minutes per epoch when both employ binomial distribution based feedback). 
After $130$ epochs, the WTM had a steady training accuracy of $100\%$, whereas the TM peaked at $99.86\%$. The WTM also had a higher testing accuracy, improving peak accuracy from $98.27\%$ to $98.63\%$. Table~\ref{tab:mnist-acc} contains test accuracy statistics of the WTM, collected from the last $50$ of $300$ epochs of single-run training (multiple runs behave similarly). As seen, the WTM learning is relatively stable, with a small difference between minimum and maximum test accuracy.


\pgfmathsetmacro\class{9}

\begin{figure}[!ht]
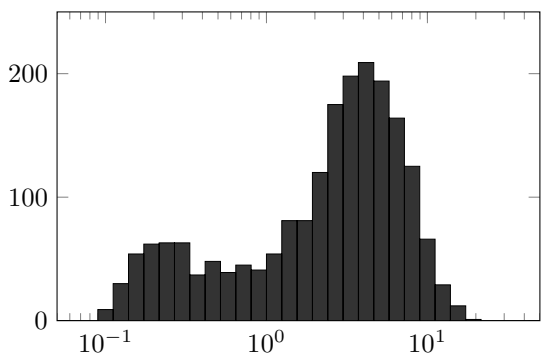

\centering
\def\hwidth{0.45\textwidth}%
\def\hheight{0.32\textwidth}%
\pgfmathsetmacro\maxcount{250}%
\pgfmathsetmacro\minweight{.05}%
\pgfmathsetmacro\maxweight{50}%
\pgfmathtruncatemacro\pcolumn{2*\class+1}%
\pgfplotstableread{mnist-c040-p080-t100-g100.w.histogram}\datatable
\histogramit{\pcolumn}{}{}\\
\caption{The distribution of weights learnt from MNIST with a WTM possessing $4{,}000$ clauses per class}
\label{fig:coefficients}
\end{figure}

Lastly, we investigated the behaviour of the WTM weight learning. In Table~\ref{tab:coefficients}, the maximum, minimum, and mean of the weights are reported per class. As seen, how the weights are distributed varies from class to class. The weights span from $0.09$ to $21.01$, and we observe that every class both have fractional and large weight clauses. As a representative example, Figure~\ref{fig:coefficients} shows the distribution of positive weights after training epoch $300$ for the digit '\class' class. The distribution reveals that the weights are relatively diverse, including two distinct peaks, around $0.6$ and $5.0$.

The benefits of weighed clauses thus becomes apparent. The clause with the largest weight has $\frac{21.01}{0.09} = 233.44$ times more impact than the clause with the smallest weight. To achieve this effect with a standard TM, one would need approximately $233$ times more copies of the first clause, compared to the second clause. Hence, the power of the WTM scheme!

\begin{table*}[!ht]
\newcounter{col}%
\def\and{&\hfil}
\pgfplotstableread{mnist-c040-p080-t100-g100.w.statistics}\dt
\centering
\begin{tabular}{l*{10}{p{2.6em}}}
\toprule
\bfseries Class&\hfil\bfseries0&\hfil\bfseries1&\hfil\bfseries2&\hfil\bfseries3&\hfil\bfseries4&\hfil\bfseries5&\hfil\bfseries6&\hfil\bfseries7&\hfil\bfseries8&\hfil\bfseries9\\
\midrule[\heavyrulewidth]
\bfseries Maximum
\setcounter{col}{0}%
\while{\thecol<10}{%
    \expandafter\and
    \pgfplotstablegetelem{1}{\thecol}\of{\dt}%
    \pgfmathprintnumberto[fixed, precision=2]{\pgfplotsretval}{\theret}%
    \theret
    \stepcounter{col}%
  }%
\\[.5ex]
\bfseries Minimum
\setcounter{col}{0}%
\while{\thecol<10}{%
    \expandafter\and
    \pgfplotstablegetelem{0}{\thecol}\of{\dt}%
    \pgfmathprintnumberto[fixed, precision=2]{\pgfplotsretval}{\theret}%
    \theret
    \stepcounter{col}%
  }%
\\
\midrule
\bfseries Mean
\setcounter{col}{0}%
\while{\thecol<10}{%
    \expandafter\and
    \pgfplotstablegetelem{2}{\thecol}\of{\dt}%
    \pgfmathprintnumberto[fixed, precision=2]{\pgfplotsretval}{\theret}%
    \theret
    \stepcounter{col}%
  }%
\\[.5ex]
\bfseries Ratio
\setcounter{col}{0}%
\while{\thecol<10}{%
    \expandafter\and
    \pgfplotstablegetelem{1}{\thecol}\of{\dt}%
    \pgfmathsetmacro\mx\pgfplotsretval
    \pgfplotstablegetelem{0}{\thecol}\of{\dt}%
    \pgfmathsetmacro\mn\pgfplotsretval
    \pgfmathsetmacro\theratio{\mx/\mn}%
    \pgfmathprintnumberto[fixed, precision=0]{\theratio}{\theret}%
    \theret
    \stepcounter{col}%
  }%
\\
\bottomrule
\end{tabular}
\caption{Clause weight statistics per class for a WTM with $4{,}000$ clauses, trained on MNIST}
\label{tab:coefficients}
\end{table*}


\subsubsection{Comparison with Other Methods}
Table~\ref{tab:comp} contrasts the WTM mean test accuracy on MNIST against vanilla versions of selected well-known classification methods, without applying performance enhancing techniques (such as boosting, ensembles, and convolution), and without data augmentation (such as deskewing, blurring, shifting, and resizing).\footnote{The results are obtained from \url{http://yann.lecun.com/exdb/mnist/} 
and the original TM paper \cite{granmo2018tsetlin}.} Note that the TM and WTM operate on the threshold-based binary encoding of the images (compression factor 8), rather than the original greyscale images. As seen, the WTM performs favourably compared to the other techniques, under comparable conditions.


\begin{table}[!ht]
\centering
\begin{tabular}{rl}
\toprule
\bfseries Method&\bfseries Accuracy\,(\%)\!\!\!\\
\midrule[\heavyrulewidth]
2-layer NN, 800 Hidden Units&98.6\\
\bfseries Weighted Tsetlin Machine&\boldmath$98.5\pm0.0$\\
K-nearest Neighbors, L3&97.2\\
3-layer NN, 500+150 Hidden Units&97.1\\
40 PCA + Quadratic Classifier&96.7\\
1,000 RBF + Linear Classifier&96.4\\
Logistic Regression&91.5\\
Linear Classifier (1-layer NN)&88.0\\
Decision Tree&87.8\\
Multinomial Naive Bayes&83.2\\
\bottomrule
\end{tabular}
\caption{Comparison of vanilla methods on MNIST}
\label{tab:comp}
\end{table}

\subsection{IMDb Sentiment Analysis}
The IMDb sentiment analysis dataset consists of $50{,}000$ highly polar movie reviews, organized into two categories: positive and negative reviews. The dataset is split evenly into $25{,}000$ reviews for training and $25{,}000$ reviews for testing. In this experiment, we use the unigrams and bigrams of each review as feature candidates, selecting the best $5{,}000$ ones based on Chi-square feature selection. Each review is thus represented by a binary feature vector of size $5{,}000$, marking the presence and absence of unigrams/bigrams.

\begin{table}[!ht]
\centering
\tableit
{89.86}{89.73}{89.80}{0.05}{89.73}{89.86}
{90.37}{90.00}{90.25}{0.16}{90.00}{90.37}
\caption{IMDb test accuracy statistics for a WTM and a TM with $25{,}000$ clauses per class}
\label{tab:imdb}
\end{table}

We first compare a WTM with $3{,}200$ class-clauses with a TM with $10{,}000$ class-clauses. After 30 epochs of training, the WTM reached a peak testing accuracy of $89.54$\%, compared to $89.38$\% for the TM.
With an equal number of clauses, $25{,}000$ per class, the WTM obtained an accuracy of $90.25\%\pm0.16\%$, significantly higher than what is achieved with the TM (Table~\ref{tab:imdb}). The average processing time per epoch was 16 minutes for the TM and 14 minutes for the WTM when both use binomial distribution based feedback.



\subsection{Connect-4 Winner Prediction}
\label{section:connect-4}
Connect-4 is a two-player board game similar to Tic-Tac-Toe, played on a $6\times7$ grid. Each player tries to connect four of her pieces in a row, either horizontally, vertically, or diagonally. The Connect-4 dataset \cite{Dua:2019} contains all $67{,}557$ legal positions of the board after eight moves, in which neither player has won yet and the next move is not forced. Each position is labelled as either a win, a loss, or a draw --- the outcome of the game after optimal play from the respective position. We assign $10\%$ of the data to testing, keeping $90\%$ for training. We further represent the board state as a binary feature vector of size $6\times7\times2$. The first $6\times7$ bits capture the placement of player one pieces, and the last $6\times7$ bits capture the placements of player two.

Using a WTM with $200$ clauses, we achieved a similar mean test accuracy to a TM with $10{,}000$ clauses over the last $200$ epochs of $1{,}000$ epochs of training; that is, 
the WTM provided similar accuracy with $50$ times fewer clauses! Table~\ref{tab:connect4} contains the resulting test accuracy statistics of a WTM and a TM with the same number of $25{,}000$ clauses per class, again, demonstrating superior performance of the WTM compared to the TM. Here, the TM spent on average 105 seconds per epoch, while the WTM used 76 seconds, which again is slightly faster despite the addition of weights.


\begin{table}[!ht]
\centering
\tableit
{82.93}{81.98}{82.49}{0.34}{82.19}{82.80}
{87.91}{86.75}{87.32}{0.20}{86.97}{87.66}
\caption{Connect-4 test accuracy statistics for a WTM and a TM with $25{,}000$ clauses per class
}
\label{tab:connect4}
\end{table}


%

\section{Conclusion and Further Work}\label{sec:conclusion}
In this paper, we introduced the Weighted Tsetlin Machine (WTM), generalizing the TM by adding weighted clauses. Instead of repeating a clause multiple times to increase its impact, we assign a weight to each clause. The weights control the impact of the clauses, and are learnt using a novel variant of Type~I and Type~II feedback. In this manner, the WTM can obtain a more compact representation of patterns. In addition, since the weights can take fractional values, the clauses can more easily be fine-tuned to represent complex patterns. We further proposed a new sampling approach for Type~I feedback generation, replacing sampling from a Bernoulli process with sampling from a binomial distribution followed by uniform sampling.

Our empirical results showed that the WTM performs comparably to the WTM when equipped with much fewer clauses (from $3$ to $50$ times less clauses). Furthermore, with the same amount of clauses, the WTM outperformed the TM accuracy-wise on MNIST, IMDb, and Connect-4.

In our further work, we intend to investigate whether the WTM approach also can improve the Regression Tsetlin Machine \cite{abeyrathna2019regressiontsetlin} for logic regression~\cite{ruczinski2003logic}, being capable of more fine-grained tuning of clauses. Another improvement, also investigated in \cite{abeyrathna2019scheme} for the regular TM, would be to introduce real-valued instead of binary inputs. This, for instance, would enable the WTM to operate directly on greyscale images, eliminating the need for binarization.

%
%
\bibliographystyle{aaai}
\bibliography{references-StarAI20}
\end{document}